\documentclass[runningheads]{llncs}
\usepackage{graphicx}
\usepackage[colorlinks]{hyperref}
\usepackage{array}
\usepackage{arydshln}
\usepackage{diagbox}
\usepackage{amsfonts}
\newcolumntype{P}[1]{>{\centering\arraybackslash}p{#1}}
\newcolumntype{M}[1]{>{\centering\arraybackslash}m{#1}}

\usepackage{amsmath}
\usepackage[final]{changes}
\usepackage[disable]{todonotes}
\definechangesauthor[name={Per cusse}, color=blue]{haofu}
\newcommand{\rhaofu}{\replaced}
\newcommand{\chaofu}{\todo}

\begin{document}
\title{Patch Transformer for Multi-tagging Whole Slide Histopathology Images}
\author{Weijian Li\inst{1} 
\and Viet-Duy Nguyen\inst{1}
\and Haofu Liao\inst{1} 
\and \\ Matt Wilder\inst{2} 
\and Ke Cheng\inst{2}
\and Jiebo Luo\inst{1}}

\institute{Department of Computer Science, University of Rochester
\and HistoWiz Inc., 760 Parkside Ave, Brooklyn, NY 11226\\
\email{\{wli69,hliao6,jluo\}@cs.rochester.edu}, \email{vnguy14@u.rochester.edu}, \\
\email{ke@histowiz.com}
}

\maketitle              
\begin{abstract}
Automated whole slide image (WSI) tagging has become a growing demand due to the increasing volume and diversity of WSIs collected nowadays in histopathology. Various methods have been studied to classify WSIs with single tags but none of them focuses on labeling WSIs with multiple tags. To this end, we propose a novel end-to-end trainable deep neural network named  \textit{Patch Transformer} which can effectively predict multiple slide-level tags from WSI patches based on both the correlations and the uniqueness between the tags. Specifically, the proposed method learns patch characteristics considering 1) patch-wise relations through a patch transformation module and 2) tag-wise uniqueness for each tagging task through a multi-tag attention module. Extensive experiments on a large and diverse dataset consisting of 4,920 WSIs prove the effectiveness of the proposed model.
\end{abstract}
\section{Introduction}
Whole slide images (WSIs) contain rich information about the morphological and functional characteristics of biological systems, which facilitate clinical diagnosis and research~\cite{wang2012managing}. To better represent image contents, pathologists frequently examine and correct the attribute tags that are inconsistent or missing for the collected WSIs. However, this tag assignment process is time-consuming and can be biased by subjective judgments, making it essential to automatically and accurately assign tags to these digital histopathology images.

To achieve this goal, several patch-based methods have been proposed for automated WSI tagging. For example, previous work~\cite{babaie2017classification,zeng2015deep} proposes to use convolutional neural networks(CNNs) to classify and retrieve WSIs with patch-level information. Hou et al.~\cite{hou2016patch} also introduce several patch-based deep models but to address a slide-level WSI classification task with a novel two-steps learning schema. A more recent work by Mercan et al.~\cite{mercan2018multi} investigate a multi-instance based model on a multiple label classification task. Their model extracts handcrafted features from prelocated Regions of Interest(ROIs) and learns slide-level labels with weakly supervised learning. The results of these studies indicate a great benefit of integrating detailed patch contents for slide-level decision making. However, their integration approaches are preliminary and constrained. 

A better way to integrate patch-level information and to automatically locate ROIs is to leverage an attention mechanism that considers the importance of each patch. Ilse et al.~\cite{ilse2018attention} propose an attention mechanism under a multi-instance learning to highlight the patch instances that contribute the most to slide-level colon cancer classification. Li et al.~\cite{li2018graph} propose a graph convolutional network (GCN) based method to learn attention weights for each patch. However, their method requires a large number of patch nodes and detailed graph structure knowledge to construct a complete graph representation for effective GCN training. Therefore, a method that can adaptively learn slide-level representations with limited prior knowledge is needed. 

One of the options is to adopt the ``Scaled Dot-Product Attention'' (SDPA), which is a self-attention mechanism introduced in the Transformer model~\cite{vaswani2017attention} for Neural Machine Translation.
\chaofu{It is abrupt to mention SDPA here. what is the benefit of SDPA? and what is its connection to your proposed method?}
It constructs rich instance-level representations considering the pairwise relationships of all given instances without higher-level structural knowledge. However, SDPA may not be the best choice considering different instance contexts between words and WSI patches. The differences in tasks themselves should also lead to different attention designs. Therefore, it is necessary to investigate and construct an appropriate attention mechanism to extract informative patch features for WSI tagging.

In summary, our contributions are as follows: 1) A novel patch based deep model \textit{Patch Transformer} is designed for multi-tagging whole slide images. To the best of our knowledge, \textit{this is the first multi-tagging approach for WSIs.} The proposed model is trained end-to-end under a multi-task learning scheme where each task is a multi-class classification problem. 2) A \textit{Patch Transformation Module} extracts patch characteristics considering global context with limited prior structural knowledge through a multi-head attention mechanism. 3) A \textit{Multi-Tag Attention Module} constructs tag-specific representations by aggregating weighted patch features. 4) Extensive experiments on a large and diverse dataset containing 4,920 WSIs demonstrate the improved performance of the proposed model compared to the state-of-the-art methods.

\section{Methods}
Given a WSI dataset $I=\{I_1, ..., I_n\}$ where each $I_n$ is a whole slide image,  we have bags of patches $B=\{B_1, ..., B_n\}$ where each $B_n$ is a bag of $M$ sampled patches from the non-background regions of $I_n$. We have $K$ sets of tags $C_k$ each has multiple classes. For the whole slide images $I$, their corresponding $k$th tags can be represented as $L_k=\{L_{k1}, ..., L_{kn}\}$, where $L_{kn}\in C_k$. Our goal is to correctly assign all $K$ tags for each image.

\begin{figure*}[t!]
	\centering
	\includegraphics[width=1.0\textwidth]{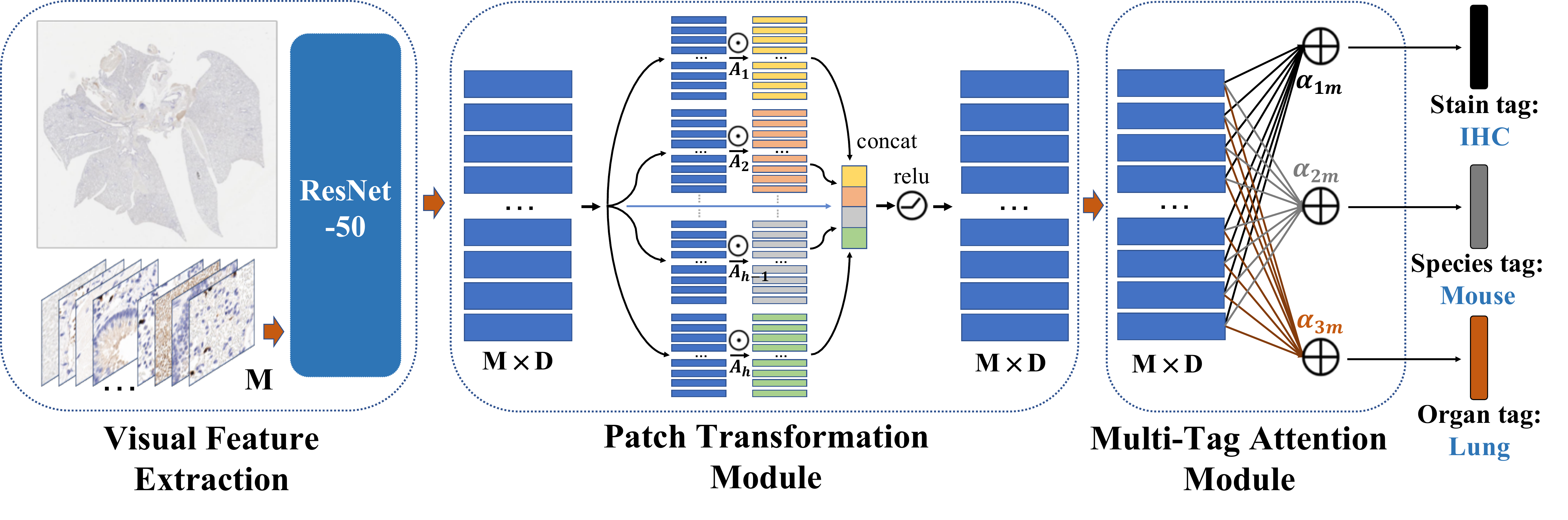}
	\vspace{-6mm}
	\caption{Overview of the proposed Patch Transformer. The proposed model mainly consists of three parts:~(1)~{\it Visual Feature Extraction}:~capturing the visual feature of each patch from the original WSI; (2)~{\it Patch Transformation Module}: producing the characteristic enhanced patch representations by an attention aggregation mechanism; (3)~{\it Multi-Tag Attention Module}: constructing tag-related global slide representations for final prediction leveraging the extracted patch features following the same attention format as (2).}
	\label{fig:framework}
	\vspace{-4mm}
\end{figure*}

\subsubsection{Patch Transformation Module} Inspired by the Transformer model~\cite{vaswani2017attention}, we introduce a patch transformation module to effectively learn patch characteristics by considering global patch contexts. As depicted in Figure~\ref{fig:framework}, the proposed module takes as inputs from the latent visual embeddings extracted by a ResNet-50 \cite{he2016deep} network. It then maps each patch feature into different attention domains through multi-head attentions. The final feature outputs are the aggregation of the obtained representations. Concretely, for the visual embedding $V\in\mathbb{R}^ {M \times D}$, the output of the module can be represented by
\chaofu{brackets already mean concatenation, what is the dimension of $f_h/head_h$? Please correct/clarify the math here. Also, the subscript h for W seems unnecessary or for different head W has different dimension?}
\begin{equation}
	\label{eq:Vprime}
	\rhaofu{
        V' = \sigma(V + W^T [\mathbf{f}_1, ..., \mathbf{f}_h]), V'\in\mathbb{R}^ {M \times D},  W \in\mathbb{R}^ {(h \times D) \times D}
	}{
	    V' = \sigma(V + W_h^T \times Concat([head_1, ..., head_h]) ), V'\in\mathbb{R}^ {M \times D},  W_h \in\mathbb{R}^ {(h \times D) \times D}
	}
\end{equation}
where $h$ represents the $h$th head in the module, $\sigma$($\cdot$) is the ReLU non-linear activation function. Each $\mathbf{f}_h$ is a feature extracted by an attention unit which we will detail later.
\rhaofu{}{
}
Different from the original task~\cite{vaswani2017attention}, here different patches share the same attribute tags. Selecting the most informative patches that contribute the most \rhaofu{to}{for} the slide level prediction becomes the main challenge. To \rhaofu{address}{approach} this issue, we \rhaofu{formulate}{adopt} the multi-head computation as an attention aggregation process to obtain characteristic-enhanced feature representations for \rhaofu{}{the later} informative patch selection. The original ``Scaled Dot-Product Attention''~\cite{vaswani2017attention} is not appropriate due to its designed patch-wise feature mix and fusion property which would diminish unique characteristics.
Instead, for the extracted patch feature matrix $V$, we \rhaofu{perform}{do} element-wise multiplication between $V$ and multi-head patch attention matrices\rhaofu{, i.e., $\mathbf{f}_h = V \odot A_h$}{: $Attention_h(V) = V \odot A_h$}, $V, A_h \in\mathbb{R}^ {M \times D}$. The attention matrix $A_h$ can be written as \rhaofu{$A_h = [\mathbf{a}_{h}, \mathbf{a}_{h}, ..., \mathbf{a}_{h}]$}{$A_h = [\overrightarrow{a_{h}}|\overrightarrow{a_{h}}|...|\overrightarrow{a_{h}}]$}, where each column \rhaofu{$\mathbf{a}_h \in\mathbb{R}^ {M \times 1}$}{$\overrightarrow{a_{h}} \in\mathbb{R}^ {M \times 1}$} is a duplicate of the attention vector for the patch features. Each weight in ${\mathbf{a}_h}$ is calculated by
\begin{equation}\label{eq:a_hn}
\rhaofu{
    a_{hm}=\text{Softmax}(W_h^T\tanh(U_h^Tv_{m})), W_h\in\mathbb{R}^{D'\times 1}, U_h\in\mathbb{R}^{D\times D'}, v_n\in\mathbb{R}^{D\times 1}
}{
    a_{hm}=Softmax(W_h^Ttanh(U_h^Tv_{m}))), W_h\in\mathbb{R}^{D'\times 1}, U_h\in\mathbb{R}^{D\times D'}, v_n\in\mathbb{R}^{D\times 1}
}
\end{equation}
We adopt a similar attention \rhaofu{mechanism to the one}{form} proposed in~\cite{ilse2018attention} for effective patch selection. \rhaofu{But we do not follow the conventional attention mechanisms by reducing the feature matrix to a unified vector}{It is worth noting that no feature fusion step, i.e., summation operation, is conducted to reduce the feature matrix to a unified vector representation as most common attention mechanisms would follow}. Instead, we add a residual connection from the input to the output and directly aggregate it with each patch's weighted representations \rhaofu{(Eq. \ref{eq:Vprime})}{through a multi-head concatenation operation}. \rhaofu{In this way}{By doing this}, we explicitly expose each patch feature's distinct characteristics while at the same time preserving its original feature representation. The final output is obtained by applying the ReLU non-linear activation function.

\subsubsection{Multi-Tag Attention Module} \rhaofu{}{So far, we have collected each patch's aggregated characteristic feature.} Our goal is to extract the most informative slide-level features for tag classifications. Based on our observation, there exists correlations among different tags, e.g. most Zebrafish slides are H\&E stained, which makes multi-task learning an appropriate approach. Meanwhile, different tags focus not only on common regions but also on tag-specific regions which results in different potential ROIs. To learn each tag's ROI adaptively and to form tag-related slide level representations, we propose a multi-tag attention module. The proposed module adopts the same attention mechanism as used in the previously introduced patch transformation module except that the output is obtained by aggregating weighted patch features. This consistent design helps our model leverage the previously learned patch characteristics to assign tag-related weights. Formally, the tag-specific representations can be represented by
\rhaofu{
    $t_k=\sum_{m=1}^{M}\alpha_{km} \times v_m'$
}
{
    \begin{equation}\label{eq:t_k}
    t_k=Attention'_k(V')=\sum_{m=1}^{M}\alpha_{km} \times v_m'
    \end{equation}
}
where $k$ represents the $k$th tag. $v_n'\in\mathbb{R}^{D'\times 1}$ is a patch feature in $V'$, $\alpha_{kn} \in\mathbb{R}^{1\times 1}$ is a tag-related weight scalar and is computed following the same format in equation (\ref{eq:a_hn}). The prediction probability for each tag of each WSI is computed by
\begin{equation}\label{eq:l_k}
\rhaofu{
    \hat{l}_k=\text{Softmax}(W_k^Tt_k), W_k\in\mathbb{R}^{D'\times D_k}, t_k\in\mathbb{R}^{D'\times 1}
}{
    \hat{l_k}=Softmax(W_k^T \times t_k), W_k\in\mathbb{R}^{D'\times D_k}, t_k\in\mathbb{R}^{D'\times 1}   
}
\end{equation}
where $D_k$ equals to the number of classes in $C_k$. The model is end-to-end trained with a combination of multi-class cross entropy losses (CE) for each tag weighted by $\lambda_k$
\begin{equation}
\rhaofu{
    \mathcal{L}=\sum_{k=1}^{K}\lambda_k\mathcal{L}_k=\sum_{k=1}^{K}\lambda_k\frac{1}{N}\sum_{n=1}^{N}\text{CE}(l_{kn},\hat{l}_{kn})
}{
    \mathbb{L}=\sum_{k=1}^{K}\lambda_k\mathbb{L}_k=\sum_{k=1}^{K}\lambda_k\frac{1}{N}\sum_{n=1}^{N}CE(l_{kn},\hat{l_{kn}})
}
\end{equation}

\begin{table}[t!]
    \centering
    \caption{Tag distribution of the dataset. }\label{tb:stat}
    \resizebox{0.96\linewidth}{!}{
    \begin{tabular}{|M{1.3cm}|M{1.2cm}M{1.2cm}M{1.2cm}M{1.0cm}M{1.3cm}M{1.3cm}M{1.0cm}M{1.8cm}|}
        \hline
        \textbf{Stain} & H\&E & IHC & Special & & & & & \\
        Count & 2803 & 1672 & 445 & & & & & \\
        \hline
        \textbf{Species} & Human & Monkey & Mouse & Pig & Rat & Zebrafish & & \\
        Count & 816 & 29 & 3435 & 33 & 439 & 168 & & \\
        \hline
        \textbf{Organ} & Bone & Brain & Breast & Cecum & Colon & Heart & Skin & Skin Dorsal\\
        Count & 89 & 238 & 264 & 119 & 338 & 160 & 752 & 67\\ 
        \hdashline
        \textbf{Organ} & Intestine & Kidney & Liver & Lung & Pancreas & Prostate & Spleen & Skin Ventral\\
        Count & 186 & 335 & 901 & 644 & 347 & 202 & 204 & 74\\
        \hline
    \end{tabular}
}
\vspace{-4mm}
\end{table}

\section{Experiments and Results}
\subsubsection{Dataset and settings} \rhaofu{The dataset used in this study contains 4,920 WSIs provided by a histopathology service company.}{The dataset used in this study is a subset that contains 4,920 randomly sampled WSIs following the uniform distribution from a database of a histopathology service company. By doing this, we keep the same tag categories and similar tag distributions as the original database.}
\chaofu{the data collection details are unnecessary and no one cares}
On average, the size of each image file is 1.17GB.
The dataset contains three slide level tags, namely: \textit{Stain}, \textit{Species}, and \textit{Organ}. In total, there are 3 stain tags, 6 species tags and 16 organ tags. Stain tag indicates the type of dye used in the histopathology staining process. Species tag indicates the type of species that the slide comes from. Organ tag indicates the organ type the slide contains. \rhaofu{}{All other information is not included in the experiment except the three attribute tags mentioned above.} Detailed tag distribution can be found in Table \ref{tb:stat}.

We first randomly split our dataset into training and testing sets with an 8:2 ratio. Then 10 percent of the training data are randomly picked and are kept as the validation set for model and parameter tuning. For each whole slide image, we use 40x resolution and apply the widely adopted Otsu~\cite{otsu1979threshold} method on the grayscale image to remove background regions. During training, $M=32$ image patches of size $512\times512$ from the non-background regions are randomly extracted. Due to the class imbalance problem and the variety of the samples, we conduct rich data augmentation operations including \rhaofu{random cropping, left-right/bottom-up flipping, and rotating.}{random cropping, random left-right/bottom-up flipping, and random rotating the image with an angle} The final patch inputs are of size $224\times224$. The 2048 dimensional outputs of \texttt{conv5\_3} of the \texttt{ResNet\_50}~\cite{he2016deep} are used as the latent visual embeddings. \texttt{ResNet\_50} is pretrained on ImageNet~\cite{deng2009imagenet} and finetuned during the training process. The model is implemented based on Tensorflow and is trained with the Adam optimizer with $lr=0.0001$, $\beta_1=0.9$, $\beta_2=0.999$. $\lambda_1=\lambda_2=\lambda_3=1$.

\begin{figure*}[t!]
	\centering
	\includegraphics[width=1\textwidth]{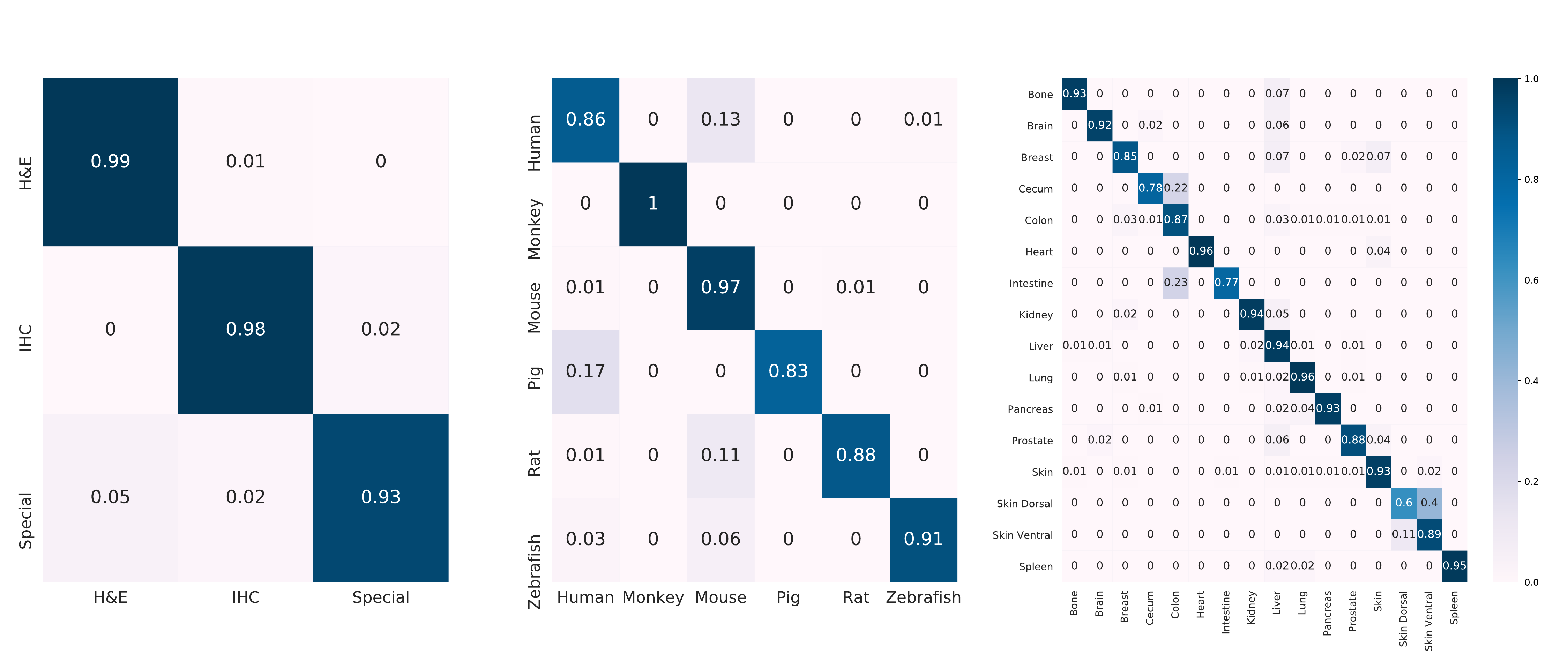}
	\vspace{-6mm}
	\caption{Normalized confusion matrices of the proposed PT-3head-MTA model for Stain tag, Species tag and Organ tag.}
	\label{fig:confusionmatrices}
	\vspace{-4mm}
\end{figure*}

\subsubsection{Comparison Methods} We compare our model with the state-of-the-art methods in: 1) whole slide image classification tasks; 2) multiple-instance learning tasks. Patch based methods~\cite{hou2016patch} follow a two-step learning process. They use the patch-level predictions for the final slide-level prediction. For GCN~\cite{kipf2016semi}, we consider WSI as a graph where each patch is a node. Graph edges are defined based on patch-to-patch spatial distance. Maxpooling is adopted as instance aggregation method. For DeepMIL~\cite{wu2015deep}, we adopt the structure that only contains visual features. To examine the effect of patch-level labels, we adopt the TwoBranches~\cite{das2018multiple} model where the additional branch targets patch-level predictions. Each patch has the same labels as the original WSI. The total loss is a combination of patch-level and slide-level cross-entropy losses. We adopt the publicly available implementation of GCN~\cite{kipf2016semi} \footnote{https://github.com/tkipf/gcn} and reimplement the other baseline models.

For fair comparison, we keep all patch extraction and prepossessing steps the same. All models adopt the same pretrained ResNet-50 structure as feature extractor. If the original model aims at single tag classification, we append additional tag classification heads to obtain a model with the same structure as used in our model. To further investigate the effect of the two proposed modules, we conduct several ablation studies.

\begin{table*}[!t]
\vspace{-2mm}
\centering
\caption{Quantitative results. PT: the proposed Patch Transformation module; MTA: the proposed Multi-Tag Attention module; SDPA: replacing our attention mechanism with the ``Scaled Dot-Product Attention'' introduced in~\cite{vaswani2017attention}. }\label{tb:quantitative}
\resizebox{0.8\linewidth}{!}{
    \begin{tabular}{l|M{1.0cm}M{1.0cm}M{1.0cm}M{1.0cm}|M{1.0cm}M{1.0cm}M{1.0cm}M{1.0cm}}
        \hline
        \textbf{Model} & 
        \multicolumn{4}{c|}{\textbf{Macro F1}} & \multicolumn{4}{c}{\textbf{Micro F1}}\\
                                                    & Stain & Species & Organ & Avg. & Stain & Species & Organ & Avg.\\
        \hline
        Patchbased-LR~\cite{hou2016patch}           & 0.937 & 0.477 & 0.378 & 0.597 & 0.972 & 0.857 & 0.556 & 0.795 \\
        Patchbased-SVM~\cite{hou2016patch}          & 0.951 & 0.754 & 0.371 & 0.692 & 0.975 & \textbf{0.951} & 0.531 & 0.819 \\
        GCN~\cite{kipf2016semi}                     & 0.961 & 0.832 & 0.822 & 0.872 & 0.981 & 0.921 & 0.881 & 0.927 \\
        DeepMIL~\cite{wu2015deep}          & 0.962 & 0.863 & 0.850 & 0.892 & 0.981 & 0.932 & 0.904 & 0.939 \\
        TwoBranches~\cite{das2018multiple}          & \textbf{0.976} & 0.845 & 0.866 & 0.895 & \textbf{0.988} & 0.932 & 0.908 & 0.942 \\
        \hline
        MTA                                         & 0.961 & 0.836 & 0.824 & 0.873 & 0.981 & 0.932 & 0.879 & 0.931\\
        PT-1head                                    & 0.957 & 0.848 & 0.850 & 0.885 & 0.981 & 0.933 & 0.904 & 0.939\\
        PT-1head-MTA                                & 0.962 & 0.846 & 0.872 & 0.893 & 0.982 & 0.933 & 0.908 & 0.941 \\
        PT-3head(SDPA)-MTA                          & 0.951 & 0.830 & 0.866 & 0.882 & 0.975 & 0.916 & 0.901 & 0.930 \\
        PT-3head-MTA                                & 0.962 & \textbf{0.889} & \textbf{0.879} & \textbf{0.910} & 0.982 & 0.939 & \textbf{0.912} & \textbf{0.944}\\
        \hline
    \end{tabular}%
}
\label{tab:result}%
\vspace{-3mm}
\end{table*}%

\subsubsection{Quantitative Results} Our quantitative evaluation results can be found in Table \ref{tb:quantitative}. Considering the class imbalance distribution of our dataset, we adopt both Macro F1 and Micro F1 scores as our evaluation metrics. For Macro F1 score, the final result is calculated by the class average values of Precision and Recall. Thus, Macro F1 score indicates an unweighted average result over all classes. It shows how our model performs in each class under the tags. For Micro F1 score, the final result is calculated based on overall predictions without considering the class categories. It shows how our model performs over the entire dataset. The confusion matrices of our best model PT-3head-MTA for the three tags are also computed and can be found in Figure~\ref{fig:confusionmatrices}. In general, our model accurately assigns the three tags to the WSIs to prove its effectiveness.

As is shown in Table \ref{tab:result}, our model outperforms most previous methods for all three tagging tasks on both Macro F1 and Micro F1 metrics. The proposed full model under a three-heads setting leads to the best overall average performance. Comparing to the two-step based models~\cite{hou2016patch}, the other models have relatively better and more stable performance across three tasks which indicates the benefit of learning the patch features jointly. GCN~\cite{kipf2016semi} and PT-3head(SDPA)-MTA~\cite{vaswani2017attention} achieve higher scores than patch-based methods but lower scores than TwoBranches and our proposed models. We consider the reason is the degradation of patch characteristics by the weighted combination of the other patches. This can be seen more clearly by comparing PT-3head(SDPA)-MTA with PT-3head-MTA where the proposed attention module is replaced in PT-3head(SDPA)-MTA with the ``Scaled Dot-Product Attention''. The TwoBranches~\cite{das2018multiple} model adopts both slide level and patch level classification losses and thus uses extra knowledge. This extra knowledge helps their model achieve the best F1 scores for stain tagging because of the similar color patterns shared by different patches. On the other hand, due to the large variance in patch textures, this strategy inhibits the performances of species and organ tagging which are more sensitive to the variations of image textures. By examining the ablations of our models, we find that combining the MTA and PT modules gives better performance than MTA or PT alone, indicating a mutual promotional effect of the two proposed modules. Furthermore, increasing the attention heads also brings additional benefit for extracting patch characteristics which results in higher F1 scores.

\begin{figure*}[t!]
	\centering
	\includegraphics[width=1\textwidth]{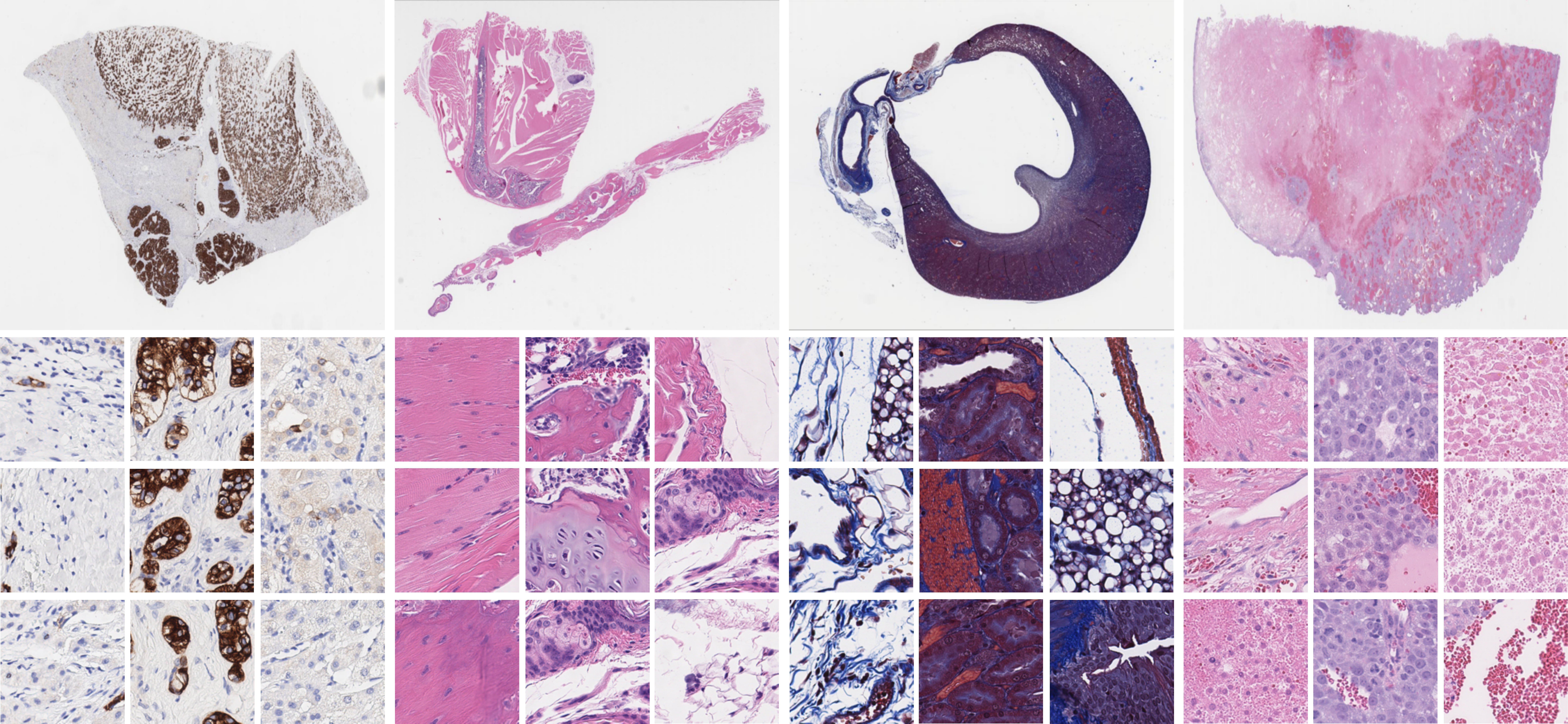}
	\vspace{-6mm}
	\caption{Multi-Tag Attention Module's attention result visualization. Each WSI has three columns. Each of the three columns belongs to one of the three tags: \textit{Stain}, \textit{Species}, and \textit{Organ} from left to right. Patches in the columns are sorted by attention weights in the corresponding tag from the top to the bottom. The ground truth tags for the four WSIs are respectively: (IHC, Human, Liver), (H\&E, Mouse, Bone), (Special Stain, Rat, Kidney), (H\&E, Mouse, Liver).}
	\label{fig:results}
	\vspace{-3mm}
\end{figure*}

\subsubsection{Qualitative Results} To visually validate the effect of multi-tag attention module, we collect and examine the attention weights as well as their corresponding patch images. Here we show four groups of examples in Figure~\ref{fig:results}. As depicted, each attention head focuses on different patch patterns. For instance, the attention head that aims at tagging stain labels has higher interest on patches with simpler textures but larger tissue areas. These patches mainly contain apparent color patterns such as the blue dots for IHC, red and pink regions for H\&E. While for the species and organ attention heads, patches with relatively complex structures and textures are assigned higher attention weights as they provide more contextual information for tagging species and organ labels. 

\section{Conclusions and Future work} 
We present a novel framework to assign multiple attribute tags for the whole slide histopathology images. Two modules are introduced, namely a patch transformation module which adopts a multi-head attention mechanism to extract and integrate patch level characteristics, and a multi-tag attention module which adaptively weights and aggregates patch features into a global slide representation for slide-level predictions targeting different tags. The proposed framework is validated on a 4,920 WSI dataset with overall improved performance over the state-of-the-art methods. More importantly, the insights on the tagging decisions can be gained effectively by visualizing the patches with the highest attention weights. Future work includes adopting extra information, e.g., multi-resolutions into the framework and developing multi-resolution fusion mechanisms to improve species and organ tag learning. Furthermore, the learned slide-level features can be explored for WSI retrieval tasks.

\subsubsection{Acknowledgement.} This work is supported in part by NSF through award IIS-1722847, NIH through the Morris K. Udall Center of Excellence in Parkinson's Disease Research, and our corporate sponsor HistoWiz.
{
    \bibliographystyle{splncs04}
    \bibliography{patchtransformer}

\begin{thebibliography}{10}
\providecommand{\url}[1]{\texttt{#1}}
\providecommand{\urlprefix}{URL }
\providecommand{\doi}[1]{https://doi.org/#1}

\bibitem{babaie2017classification}
Babaie, M., Kalra, S., Sriram, A., Mitcheltree, C., Zhu, S., Khatami, A.,
  Rahnamayan, S., Tizhoosh, H.R.: Classification and retrieval of digital
  pathology scans: A new dataset. In: CVPR-Workshops. pp. 8--16 (2017)

\bibitem{das2018multiple}
Das, K., Conjeti, S., Roy, A.G., Chatterjee, J., Sheet, D.: Multiple instance
  learning of deep convolutional neural networks for breast histopathology
  whole slide classification. In: ISBI. pp. 578--581. IEEE (2018)

\bibitem{deng2009imagenet}
Deng, J., Dong, W., Socher, R., Li, L.J., Li, K., Fei-Fei, L.: Imagenet: A
  large-scale hierarchical image database. In: CVPR. pp. 248--255. IEEE (2009)

\bibitem{he2016deep}
He, K., Zhang, X., Ren, S., Sun, J.: Deep residual learning for image
  recognition. In: CVPR. pp. 770--778 (2016)

\bibitem{hou2016patch}
Hou, L., Samaras, D., Kurc, T.M., Gao, Y., Davis, J.E., Saltz, J.H.:
  Patch-based convolutional neural network for whole slide tissue image
  classification. In: CVPR. pp. 2424--2433 (2016)

\bibitem{ilse2018attention}
Ilse, M., Tomczak, J., Welling, M.: Attention-based deep multiple instance
  learning. In: ICML. pp. 2132--2141 (2018)

\bibitem{kipf2016semi}
Kipf, T.N., Welling, M.: Semi-supervised classification with graph
  convolutional networks. arXiv preprint arXiv:1609.02907  (2016)

\bibitem{li2018graph}
Li, R., Yao, J., Zhu, X., Li, Y., Huang, J.: Graph cnn for survival analysis on
  whole slide pathological images. In: MICCAI. pp. 174--182. Springer (2018)

\bibitem{mercan2018multi}
Mercan, C., Aksoy, S., Mercan, E., Shapiro, L.G., Weaver, D.L., Elmore, J.G.:
  Multi-instance multi-label learning for multi-class classification of whole
  slide breast histopathology images. TMI  \textbf{37}(1),  316--325 (2018)

\bibitem{otsu1979threshold}
Otsu, N.: A threshold selection method from gray-level histograms. IEEE
  transactions on systems, man, and cybernetics  \textbf{9}(1),  62--66 (1979)

\bibitem{vaswani2017attention}
Vaswani, A., Shazeer, N., Parmar, N., Uszkoreit, J., Jones, L., Gomez, A.N.,
  Kaiser, {\L}., Polosukhin, I.: Attention is all you need. In: NeurIPS. pp.
  5998--6008 (2017)

\bibitem{wang2012managing}
Wang, F., Oh, T., Vergara-Niedermayr, C., Kurc, T., Saltz, J.: Managing and
  querying whole slide images. In: Proceedings of SPIE--the International
  Society for Optical Engineering. vol.~8319. NIH Public Access (2012)

\bibitem{wu2015deep}
Wu, J., Yu, Y., Huang, C., Yu, K.: Deep multiple instance learning for image
  classification and auto-annotation. In: CVPR. pp. 3460--3469 (2015)

\bibitem{zeng2015deep}
Zeng, T., Ji, S.: Deep convolutional neural networks for multi-instance
  multi-task learning. In: ICDM. pp. 579--588. IEEE (2015)

\end{thebibliography}
}

\end{document}